\documentclass[10pt,twocolumn,letterpaper]{article}
\usepackage[pagenumbers]{cvpr}              
\usepackage[accsupp]{axessibility} 
\usepackage{graphicx}
\usepackage{amsmath}
\usepackage{amssymb}
\usepackage{booktabs}
\usepackage{pifont}
\usepackage{multirow}
\usepackage[dvipsnames]{xcolor}
\usepackage[dvipsnames]{colortbl}
\definecolor{Gray}{gray}{0.9}
\definecolor{LightCyan}{rgb}{0.88,1,1}
\newcolumntype{a}{>{\columncolor{Gray}}c}
\newcolumntype{b}{>{\columncolor{white}}c}
\newcommand{\methodname}{Dr$^2$Net}

\usepackage[dvipsnames]{xcolor}

\definecolor{cvprblue}{rgb}{0.21,0.49,0.74}

\usepackage[pagebackref,breaklinks,colorlinks,citecolor=cvprblue]{hyperref}

\title{Dr$^2$Net: \underline{D}ynamic 
\underline{R}eversible \underline{D}ual-\underline{R}esidual \underline{Net}works  for \\
Memory-Efficient   Finetuning}

{\author{
Chen Zhao$^{1}$ 
\quad Shuming Liu$^{1}$
\quad Karttikeya Mangalam$^{2}$
\quad Guocheng Qian$^{1}$ \\
\quad Fatimah Zohra$^{1}$
\quad Abdulmohsen Alghannam$^{1}$
\quad Jitendra Malik$^{2}$
\quad Bernard Ghanem$^{1}$
\and
$^{1}$King Abdullah University of Science and Technology, Saudi Arabia \quad $^{2}$UC Berkeley, US\\
}}

\begin{document}
\maketitle
\begin{abstract}
Large pretrained models are increasingly crucial in modern computer vision tasks. These models are typically used in downstream tasks by end-to-end finetuning, which is highly memory-intensive for tasks with high-resolution data, e.g., video understanding, small object detection, and point cloud analysis. In this paper, we propose \textbf{D}ynamic \textbf{R}eversible  \textbf{D}ual-\textbf{R}esidual \textbf{Net}works, or \methodname{}, a novel family of network architectures that acts as a surrogate network to finetune a pretrained model with substantially reduced memory consumption. \methodname{} contains two types of residual connections, one maintaining the residual structure in the pretrained models, and the other making the network reversible. Due to its reversibility, intermediate activations, which can be reconstructed from output, are cleared from memory during training. We use two coefficients on either type of residual connections respectively, and introduce a dynamic training strategy that seamlessly transitions the pretrained model to a reversible network with much higher numerical precision. We evaluate Dr$^2$Net on various pretrained models and various tasks, and show that it can reach comparable performance to conventional finetuning but with significantly less memory usage. Code will be available at \url{https://github.com/coolbay/Dr2Net}.

\end{abstract}   
\section{Introduction}
\label{sec:intro}

Large pretrained models play an increasingly crucial role in modern computer vision tasks. These large models, such as ViTs~\cite{dosovitskiy2021image} and Swin transformers~\cite{liu2021swin,vswin}, are pretrained on large-scale datasets~\cite{deng2009imagenet,zisserman2017kinetics}, by various means such as fully-supervised learning~\cite{dosovitskiy2021image,liu2021swin},  self-supervised learning~\cite{he2022masked,tong2022videomae,feichtenhofer2022masked,oquab2023dinov2} or vision-language pretraining~\cite{radford2021learning}.  They have strong representational capacity due to the large model scale and the data scale, and therefore become indispensable for various downstream tasks~\cite{zhang2023dino,zhao2023re2tal,botach2022end,Pix4Point,soldan2021mad,grauman2022ego4d,alcazar2022end}. 

\begin{figure}[t]
\begin{center}
\footnotesize
\includegraphics[width=0.45\textwidth]{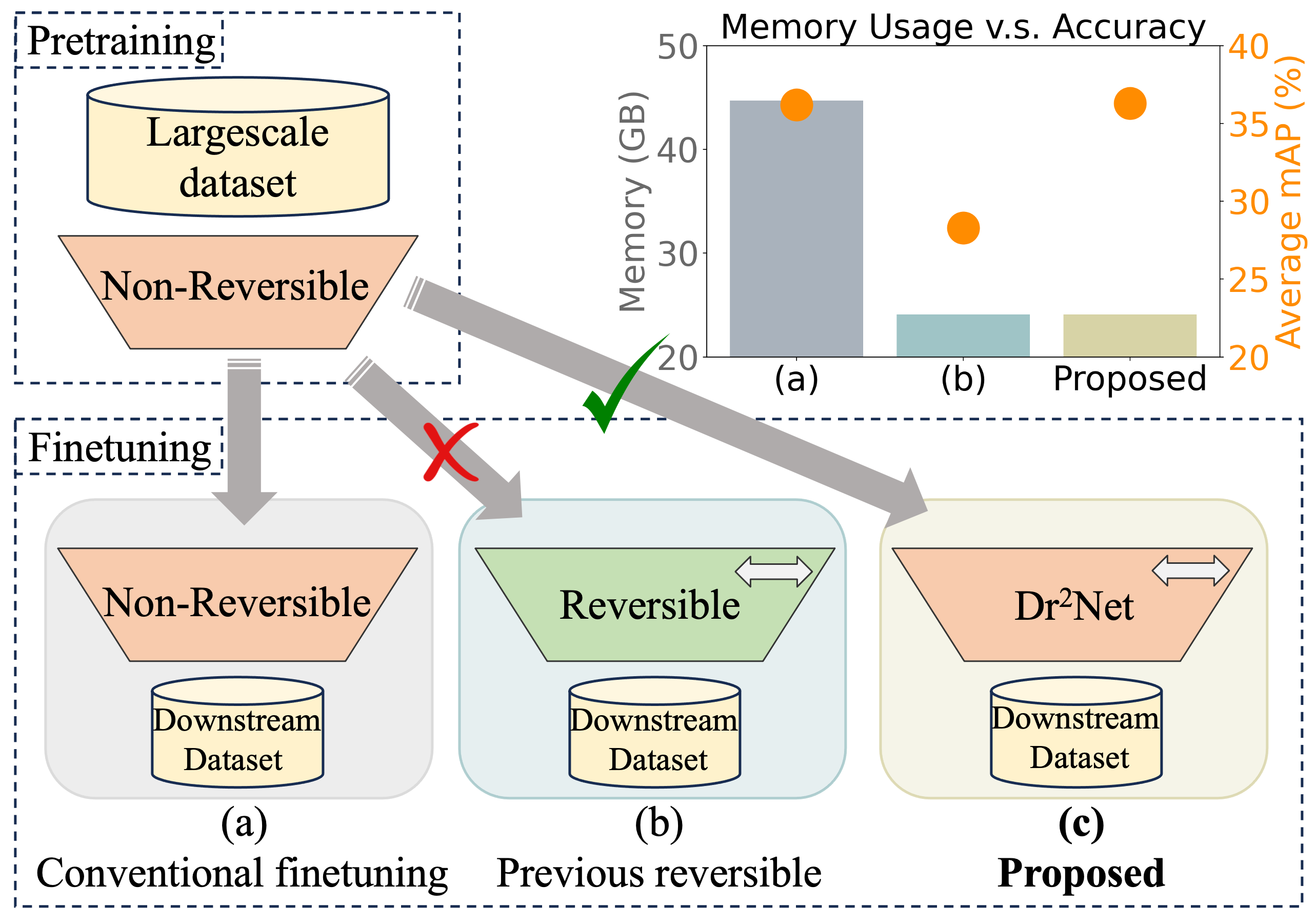}
\end{center}
\vspace{-15pt}
\caption{\textbf{Comparison of different ways of finetuning from pretrained non-reversible models. } (a) \textit{Conventional finetuning} uses the same non-reversible architecture in the downstream task, initialized with the pretrained parameters. It consumes high GPU memory. (b) \textit{Previous reversible} methods (e.g., \cite{revnet, revvit, zhao2023re2tal}) cannot finetune from pretrained non-reversible models on the downstream task due to architecture discrepancy. They show reduced accuracy when training from scratch on the downstream. (c) \textit{Our proposed \methodname{}} can directly finetune from pretrained non-reversible networks, significantly saving memory while preserving accuracy. The top-right chart illustrates memory usage and accuracy for temporal action detection on ActivityNet-v1.3~\cite{caba2015activitynet} using VSGN~\cite{zhao2021video} and Video Swin~\cite{vswin}.}
\vspace{-18pt}
\label{fig:teaser}
\end{figure}

Although these pretrained large models have shown good generality, they need to be end-to-end finetuned on specific downstream tasks to reach an optimal performance~\cite{zhang2023dino,liu2022etad, cheng2022tallformer,zhao2023re2tal,botach2022end,Pix4Point,liu2024adatad}. End-to-end finetuning refers to training the backbone, which is initialized with a pretrained model, simultaneously with the task-specific network during finetuning. For example, it is a common practice in image object detection that the backbone is initialized from a model pretrained on ImageNet classification~\cite{deng2009imagenet} and finetuned end-to-end on the object detection datasets~\cite{zhang2023dino}. For the task of video temporal action localization, recent research has shown  performance boost of using end-to-end finetuning compared to frozen-backbone finetuning~\cite{liu2022etad, cheng2022tallformer,zhao2023re2tal,liu2024adatad}. For self-supervised pretrained models such as MAE~\cite{he2022masked}, end-to-end finetuning is required to reach even a decent performance for downstream tasks.

However, end-to-end finetuning is memory intensive, especially for those large models on a task with high-dimension or high-resolution data, as shown in Fig.~\ref{fig:teaser}~(a).
For example, in long-form video understanding tasks, \eg, temporal action localization~\cite{liu2022etad, cheng2022tallformer,zhao2023re2tal}, thousands of video frames need to be processed at a time for long-term reasoning. Without dramatically downscaling the resolution, it is even impossible to finetune a Video Swin - large model with a video of 30 seconds in the largest GPU, \textit{i.e.}, A100 with 80 GB memory~\cite{zhao2023re2tal}. Therefore, reducing GPU memory consumption is a vital problem in finetuning large models.

Recently, reversible networks have demonstrated their efficacy in significantly reducing memory consumption during training~\cite{revnet, irevnet, revgraphnn, revvit, zhao2023re2tal}. They can reconstruct intermediate activations from network output, and therefore don't need to store those activations in memory during the forward process.
However, existing reversible networks~\cite{revnet, irevnet, revgraphnn, revvit} are not able to leverage pretrained models, and have to be trained from scratch, which leads to inferior performance as shown in Fig.~\ref{fig:teaser}~(b).
While the more recent work Re$^2$TAL~\cite{zhao2023re2tal} proposed a rewiring strategy enabling the reuse of pretrained model architectures and parameters, it still requires pretraining the reversible model before finetuning it on the downstream task to maintain performance. A major challenge in directly fine-tuning reversible networks from pretrained models is the inherent architectural disparity. The majority of existing pretrained models are designed as non-reversible networks, making direct transfer learning to the distinctly different architecture of reversible networks in the downstream tasks challenging.

To reduce memory consumption without compromising performance when finetuning pretrained non-reversible models on downstream tasks, in this paper, we propose a family of network architectures, dubbed as \textbf{D}ynamic \textbf{Re}versible 
\textbf{D}ual-\textbf{Re}sidual \textbf{Net}works or \methodname{}. \methodname{} acts as a surrogate backbone network during finetuning, and can be seamlessly initialized from  pretrained non-reversible models. 
\methodname{} is essentially a super network encompassing both the pretrained non-reversible architecture and the downstream reversible architecture. It employs two types of residual connections: one preserves the residual structure of the pretrained non-reversible architecture, while the other facilitates reversibility. By applying two distinct coefficients to these residual connections,
we can control the network's proximity to either architecture. 
During finetuning, we dynamically update the coefficients such that the network seamlessly transitions from the pretrained non-reversible model to a reversible network of increased numerical precision (referred to as a robust reversible network).  This design effectively bridges the architectural gap between the two types of networks.

We summarize our contributions as follows.
\begin{itemize}
    \item We propose a novel family of network architectures dubbed as Dynamic Reversible Dual-Residual Networks (\methodname{}) to finetune any pretrained model with substantially reduced memory consumption. 
     \item We introduce a dynamic finetuning strategy to seamlessly transition any network to a robust reversible network, achieving performance comparable to conventional finetuning while conserving memory. 
    \item We have shown the effectiveness of \methodname{} on various pretrained models such as Swin~\cite{liu2021swin} and ViT~\cite{dosovitskiy2021image}, and a broad range of vision tasks such as temporal action detection~\cite{xu2020g}. \methodname{} significantly reduces memory while preserving accuracy.
\end{itemize}

\section{Related Works}
\label{sec:formatting}

\subsection{Large pretrained models}

Large pretrained models~\cite{he2022masked,tong2022videomae,oquab2023dinov2,vswin}, due to their large model scales and their large-scale training  data, have demonstrated impressive performance in various computer vision tasks.  
Different pretraining mechanisms have been explored in the literature. Fully-supervised classification, e.g., image classification~\cite{liu2021swin, dosovitskiy2021image} on ImageNet~\cite{deng2009imagenet} and video action classification~\cite{slowfast,vswin} on Kinetics~\cite{zisserman2017kinetics}, is a common pretraining task when the data categories are available.
When annotation is scarce, self-supervised learning, e.g, MAE~\cite{he2022masked}, VideoMAE~\cite{tong2022videomae} and DINOv2~\cite{oquab2023dinov2}, is an effective way to leverage large-scale unlabeled data. These models can scale up more easily by utilizing the vast amount of images and videos out there without human  annotation. If paired language descriptions for the vision data are available, vision-language pretraining can be utilized, e.g., CLIP~\cite{radford2021learning} and Frozen~\cite{bain2021frozen}.

All these types of pretrained models can benefit downstream tasks through finetuning. 
It has been shown that finetuning from large pretrained models achieves significantly improved performance than training from scratch for various downstream tasks~\cite{zhang2023dino,zhao2023re2tal,botach2022end,Pix4Point,tong2022videomae}. However, the majority of existing pretrained models are non-reversible, and consume a large amount of GPU memory when used for downstream finetuning. In this paper, we propose a novel family of reversible networks for downstream finetuning, which can directly leverage these pretrained models.

\subsection{Memory-efficient training}

The computational demands, \eg GPU memory,  impede deep neural network training. Various techniques have been proposed to mitigate this issue. For instance, mixed precision training~\cite{micikevicius2017mixed} reduces the numerical precision of certain model layers while maintaining performance, thereby lowering memory usage. Another approach, activation checkpointing~\cite{chen2016training}, stores only specific intermediate activations in the forward pass and recomputes the others during backpropagation. However, their memory costs scale linearly or sublinearly  with the number of network layers, posing challenges for deeper networks. Besides these architecture-agnostic approaches, efforts to design specific memory-efficient networks have also been fruitful. An exemplary case is the reversible network~\cite{revnet,mangalam2022reversible}, which requires storing only the final feature map during forward propagation. This storage requirement remains constant irrespective of network depth. In backpropagation, the reversible network efficiently reconstructs intermediate feature maps from deep to shallow layers, offering a more scalable solution than activation checkpointing. In this paper, we adopt the idea of reversible networks for memory efficiency.

\subsection{Reversible networks}
Reversible networks originated from the idea of invertible transformations in NICE~\cite{dinh2014nice, dinh2016density},  which inspires subsequent  architectures proposed for various purposes, for example
normalizing-flow based image generation ~\cite{ho2019flow++, kingma2018glow}, signal reconstruction~\cite{liu2021invertible,mou2023ste} and memory efficiency~\cite{revnet,mangalam2022reversible,zhao2023re2tal}. RevNet~\cite{revnet} adapts the NICE transformeation for ResNets~\cite{he2016deep} and proposes a reversible backpropagation algorithm that significantly reduces the GPU training memory cost. RevViT~\cite{mangalam2022reversible} further adapts the NICE transformers to Vision Transformers~\cite{dosovitskiy2021image} and achieves performance parity across a variet of tasks. Re$^2$TAL~\cite{zhao2023re2tal} proposes a method to rewire a pretrained non-reversible backbone into a reversible backbone. But their proposed method still needs fine-tuning the reversible network on the pretraining task using the pre-training dataset. However, quite often, downstream practitioners do not have ready access to the pre-training dataset or the pre-training implementation and recipes. Further, it fails on finetuning from self-supervised learned models, such as VideoMAE~\cite{tong2022videomae}. Reversible networks are an effective approach for conserving memory, yet existing methods are not able to transfer the parameters from a non-reversible network to a reversible network. In this paper, we propose a new type of reversible networks, which enable directly finetuning from parameters of pretrained non-reversible networks in the downstream tasks.

\subsection{Memory-intensive tasks}

Many computer vision tasks, which involve  high-dimension or high-resolution data, are highly memory intensive, such as long-form video understanding, small object detection. 

\noindent\textbf{\textbf{Long-form video understanding.}} Temporal action detection (TAD)~\cite{xu2020g, lin2019bmn,zhao2023re2tal,liu2024adatad,zhao2022segtad,cheng2022tallformer,ramazanova2023owl} is a typical long-form video understanding task. It requires reasoning among a large number of video frames, and therefore uses a lot of GPU memory. We cannot even feed a video of 30 seconds into the largest GPU without significantly downscaling the video resolution. To enable training with a long sequence of video frames, most methods in the literature adopt the feature-based mechanism, where they freeze the gigantic pretrained backbone and only train the TAD-specific layers~\cite{lin2019bmn,xu2020g,zhao2021video}. However, this inevitably sacrifices accuracy. Some recent methods propose to do end-to-end training by sampling a subset of the video snippets for processing (e.g., TallFormer~\cite{cheng2022tallformer}) or for back-propagation (e.g., ETAD~\cite{liu2022etad}).  
However, these snippet-sampling based methods require each video snippet to be independently encoded, and cannot perform global temporal aggregation. 

\noindent\textbf{Object detection in large images}. To accurately detect small objects, state-of-the-art object detectors (e.g., DINO~\cite{zhang2023dino}) rely on a large-resolution input image, such as $1024\times1024$, requiring massive GPU memory for training. Consequently, limited model sizes or batch sizes can be utilized, restricting the detection accuracy.
Additionally, object detection heavily relies on the pretrained image backbone, which is utilized to initialize the detection backbone and then finetuned for the detection task. Studies show that finetuning from models pretrained on larger image classification datasets~\cite{he2022masked} or detection datasets~\cite{liu2021swin,zhang2023dino} can significantly boost the detection performance. 

In this work, we use our proposed Dr$^2$Net to dramatically reduce memory consumption of end-to-end finetuning for these memory-intensive tasks. Using the saved memory, higher input resolutions or larger models can be utilized to reach higher performance.

\section{Methodology}

\subsection{Problem formulation}

\begin{figure}[t]
\begin{center}
\footnotesize
\includegraphics[width=0.45\textwidth]{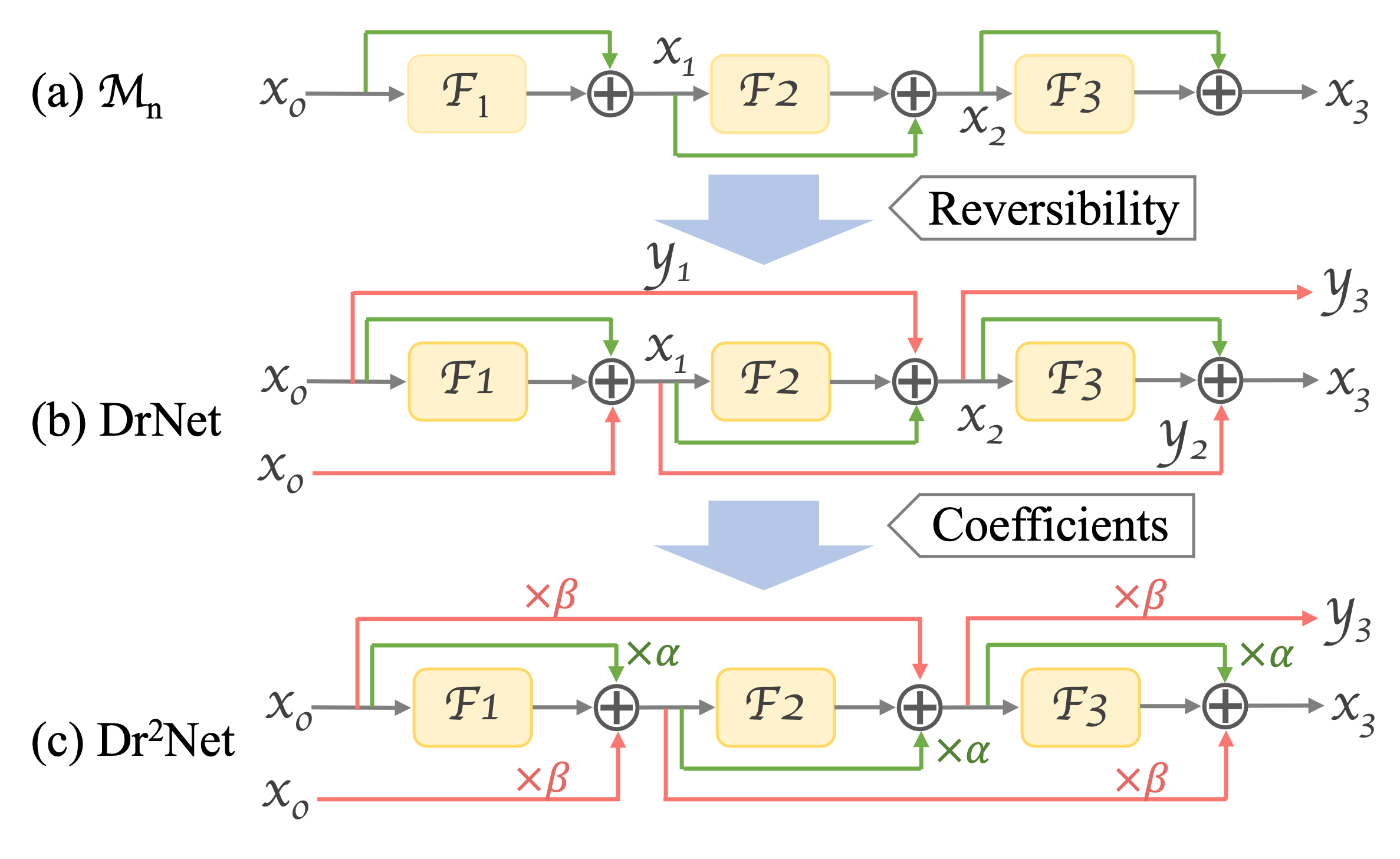}
\end{center}
\vspace{-18pt}
\caption{\textbf{Transforming a pretrained non-reversible network architecture $\mathcal{M}_n$ into our proposed \methodname{}.}  \textbf{(a) $\mathcal{M}_n$}: the pretrained non-reversible network with three blocks $\mathcal{F}_i, i=1,2,3$. Considering that most contemporary networks have residual connections, we illustrate the network with residual connections in the figure (green arrows), though our method doesn't restrict $\mathcal{M}_n$ to be residual networks. \textbf{(b) DrNet:} a reversible network obtained by adding a new group of residual connections (pink arrows) to $\mathcal{M}_n$. \textbf{(c) \methodname{}}: our proposed reversible network obtained by adding coefficients $\alpha$ and $\beta$ to the two groups of residual connections respectively. \methodname{} is equivalent to $\mathcal{M}_n$ when $\alpha=1$ and $\beta=0$.
Note that the blocks $\mathcal{F}_i$ can be of any architectures following \cite{zhao2023re2tal}, and there can be any number of $\mathcal{F}_i$ blocks in each network.  }
\vspace{-10pt}
\label{fig:unify_res_rev}
\end{figure}

Given a pretrained model, which is usually a non-reversible neural network, e.g., Video Swin~\cite{vswin} trained on Kinetics~\cite{zisserman2017kinetics}, we denote its architecture as $\mathcal{M}_n$ (Fig.~\ref{fig:unify_res_rev}~(a)), and its parameters as $\theta_n$.
Our objective is to finetune the model on a downstream task, such as video temporal action detection~\cite{zhao2023re2tal, cheng2022tallformer, liu2022etad},  in a memory-efficient manner. Typically, conventional finetuning involves transferring both the architecture and the parameters from the pretrained model to the downstream task, as illustrated in Fig.~\ref{fig:teaser}~(a).  Concretely, the same architecture $\mathcal{M}_n$ is directly utilized as the backbone architecture in the downstream task, and it is initialized with the parameter values $\theta_n$ during the finetuning process.

However, finetuning on a downstream task with high data dimension or resolution is memory-intensive. To mitigate this, instead of using the same architecture $\mathcal{M}_n$ in the downstream task, we propose to transform the pretraining architecture  $\mathcal{M}_n$ to a reversible one $\mathcal{M}_r$ as the downstream backbone, and initalize its parameters with the same values $\theta_n$ for finetuning. 
This approach raises two key questions: (1) How do we transform the architecture to a reversible one that can seamlessly reuse the parameter values $\theta_n$?  (2) How do we effectively finetune the reversible network in a memory-efficient setting? Sec.~\ref{sec:reversible_residual}  and Sec.~\ref{sec:fintuning}  will address these two questions respectively.

\subsection{Dynamic Reversible Dual-Residual Networks} \label{sec:reversible_residual}

To transform a network into a reversible one, recently, Re$^2$TAL~\cite{zhao2023re2tal} proposed to rewire the residual connections  of a non-reversible network (Fig.~\ref{fig:unify_res_rev}~(a)) to obtain a reversible architecture. During its rewiring process, though the basic blocks $\mathcal{F}_i$ are maintained,  the macro architecture has a significant change. Consequently, the obtained reversible network is mathematically a different function from the original network , and cannot be directly finetuned from the parameters $\theta_n$ of the original network on the downstream task. 

What if we can maintain the macro architecture during the network transformation?
Actually, initializing network parameters from a different network has been studied in the literature~\cite{carreira2017quo,bain2021frozen}. In I3D~\cite{carreira2017quo} and Frozen~\cite{bain2021frozen}, researchers attempted to initialize video networks from parameters of pretrained image networks. They adopted a parameter initialization mechanism that makes the video network equivalent to the  image network before finetuning by initializing the extra parameters in the video network to certain values. Inspired by these, we propose to transform the architecture with minimum modification, and ensure   equivalency between the pretrained network and the downstream network at the beginning of finetuning.

\textit{To obtain a reversible downstream network}, instead of rewiring the residual connections in the original network,  we add new residual connections, as illustrated in Fig.~\ref{fig:unify_res_rev}~(b).  Note that since most contemporary networks have residual connections, we illustrate the pretrained architecture $\mathcal{M}_n$ as a residual network following \cite{zhao2023re2tal}, though our method does not restricted $\mathcal{M}_n$ to  residual networks. At each block $\mathcal{F}_i$, we add a new residual connection, the pink arrow in Fig.~\ref{fig:unify_res_rev}~(b), that skips two blocks $\mathcal{F}_i$ and $\mathcal{F}_{i+1}$. We replicate the original input $x_0$ as the input  to the first new residual connection to form a second pathway. The obtained network is a reversible \textbf{D}ual-\textbf{R}esidual \textbf{N}etwork, DrNet for short, whose reversibility will be detailed later in this section and proved in the appendix. DrNet preserves the original residual connections. However, it still has an architectural discrepancy from the original network $\mathcal{M}_n$ due to the newly introduced residual connections (pink arrows).

\textit{To enable initializing the  reversible network to be the equivalent of the pretrained $\mathcal{M}_n$},  we introduce two coefficients on the two types of residual connections respectively. We use $\alpha \in [0,1]$  on the  original residual connections (green), and $\beta\in [0,1]$ on our newly added residual connections (pink).  With these two coefficients, we actually obtain a family of reversible networks with $\alpha$ and $\beta$ set to different values. These two coefficients can be dynamically adjusted during finetuning (see Sec.~\ref{sec:fintuning}), so we call the obtained new architecture \textbf{D}ynamic \textbf{R}eversible \textbf{D}ual-\textbf{R}esidual  \textbf{N}etworks, \methodname{} for short, as illustrated in Fig.~\ref{fig:unify_res_rev} (c).

We use \methodname{} for downstream finetuning with the parameters $\theta_n$ of the pretrained network $\mathcal{M}_n$ as initialization. When initializing \methodname{}, we make $\alpha=1$ and $\beta=0$, such that it becomes exactly the same architecture as the pretrained network $\mathcal{M}_n$. In this way, we can seamlessly initialize \methodname{} using the parameters $\theta_n$ of the pretrained network $\mathcal{M}_n$. When we make $\alpha=1$ and $\beta=1$, \methodname{} becomes DrNet above; when we make $\alpha=0$ and $\beta=1$, \methodname{} becomes an architecture as in Re$^2$TAL~\cite{zhao2023re2tal}.

We mathematically formulate the computation of the $i^{\textrm{th}}$ module in  \methodname{} as follows
\begin{equation}
\left\{\begin{array} { l }
{ y _ { i } = \beta \times x _ { i-1 }  } \\
{ x _ { i } = \mathcal{G }_i ( x_{ i-1 } ) + y_{ i-1 } },
\end{array}  \right.
\label{eq:forward_lambda}
\end{equation}
where $y_0=x_0$, and $\mathcal{G}_i( x_{ i-1 } ) = \mathcal{F}_i(x_{i-1}) +  \alpha \times x_{i-1}$.  If the pretrained network $\mathcal{M}_n$ doesn't have residual connections,  $\alpha=0$. We can observe that this \methodname{} module is reversible as long as $\beta \neq 0$. Its reverse computation is formulated as follows
\begin{equation}
\left\{\begin{array}{l}
x_{i-1}=y_{i}/\beta \\
y_{i-1}=x_{i}-\mathcal{G}_{i}(x_{i-1}).
\end{array}
\right.
\label{eq:reverse_lambda}
\end{equation} 
Recall that we need to make $\beta=0$ to make \methodname{} equivalent to the pretrained network $\mathcal{M}_n$, which conflicts the $\beta\neq0$ requirement here. We will discuss this in Sec.~\ref{sec:fintuning}. Due to the reversibility of each module, all the intermediate activations $x_i$ and $y_i$ can be reconstructed from the output, and hence  don't need to be cached in memory~\cite{zhao2023re2tal,revvit,revnet}.

\subsection{Memory-efficient finetuning} \label{sec:fintuning}

To finetune our reversible network \methodname{} in a memory-efficient manner, we clear intermediate activations from memory during the forward process, and customize the back propagation to reconstruct those activations from the output using Eq.~\ref{eq:reverse_lambda} following the implementations in \cite{revnet,revvit}. As \cite{revnet} pointed out, in reversible networks, though the activations can be exactly reconstructed when done in exact arithmetic, numerical error may be accumulated during back propagation due to floating point computation with limited precision. If the numerical error is within a certain level, it 
does not affect the performance; otherwise, training will be impaired.
In this subsection, we will discuss how to effectively finetune our \methodname{} in a memory-efficient manner with minimum influence from the numerical error.

\subsubsection{Vanilla finetuning }

As described in Sec~\ref{sec:reversible_residual}, we set $\alpha=1$ and $\beta=0$ to make \methodname{} equivalent to the pretrained network $\mathcal{M}_n$ at the beginning of the finetuning.
However, $\beta$ can not be 0 because it will be used in the denominator in the reverse process, as shown in Eq.~\ref{eq:reverse_lambda}. To circumvent this constraint, we use a small value for $\beta$ instead. During our experiments, we find that when $\beta$ is too small, the numerical error will corrupt the training. As a tradeoff between numerical precision and the resemblance of \methodname{} and $\mathcal{M}_n$, we make $\beta=0.1$.

\begin{figure}[t]
\begin{center}
\footnotesize
\includegraphics[width=0.45\textwidth]{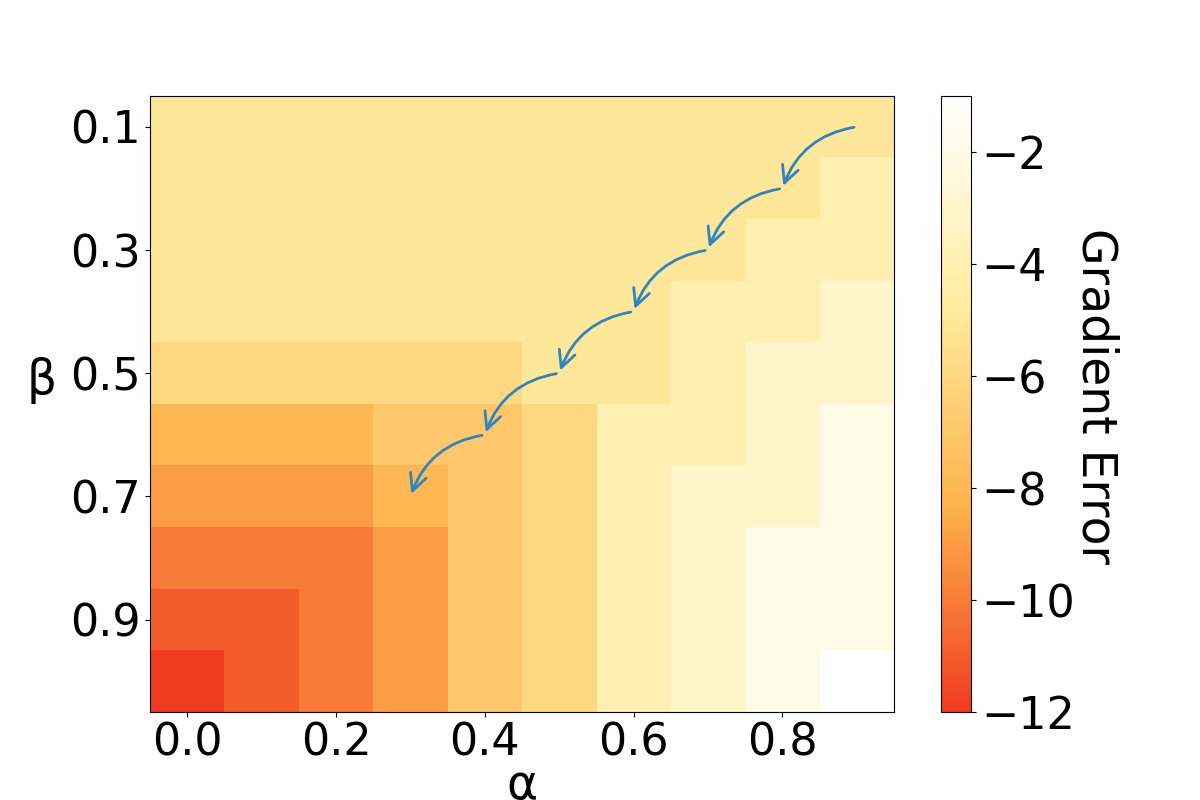}
\end{center}
\vspace{-18pt}
\caption{\textbf{Gradient error levels with different $\alpha$ and $\beta$ values.} The scales on the right colorbar represent $10^{-12} \sim 10^{-2}$. When $\alpha=1$ and $\beta=0.1$ (top-right) at the beginning of finetuning, the error level is $10^{-5}$. The errors are the smallest when $\alpha=0$ and $\beta=1$ (bottom-left) and the largest when $\alpha=1$ and $\beta=1$ (bottom-right). The error level in the middle area $10^{-7}$ is already acceptable. The blue arrows represent an ideal evolution path of the two coefficients over the finetuning process: progressively approaching the values that produce acceptable gradient errors. }
\vspace{-10pt}
\label{fig:grad_error}
\end{figure}

\textit{Can we keep using the same architecture of \methodname{} with $\alpha=1$ and $\beta=0.1$  throughout the finetuning process?} 
To answer this question, we need to know whether and how the values of $\alpha$ and $\beta$ will influence the numerical error of \methodname{} finetuning. To this end, we carry out a study on the relationship between the back propagation error and the values of $\alpha$ and $\beta$. To train \methodname{}, besides the memory-efficient training that clears intermediate activations and reconstructs them during back propagation, we can also store
its intermediate activations in GPU memory and perform common back propagation, which doesn't save memory but provides accurate gradient values not impacted by numerical errors as a reference. We compare the gradient values between the two ways of training with different $\alpha$ and $\beta$ values.  In Fig.~\ref{fig:grad_error}, we plot the gradient error levels (\ie magnitudes) of our \methodname{} with Video Swin-tiny~\cite{vswin} under FP32.

We can see that the error level is the lowest $10^{-12}$ when $\alpha=0$ and $\beta=1$ on the bottom-left corner; it is the highest $10^{-1}$ when both $\alpha$ and $\beta$ are close to 1  at the same time on the bottom-right corner. The error level in the middle area is around $10^{-7}$, which is already acceptable. Our selection of $\alpha=1$ and $\beta=0.1$ on the top-right corner to initialize the architecture has an error level of $10^{-5}$, which is not detrimental to the training, but doesn't give precise results. That indicates that if we keep using these values, it is probably very hard to reach the optimal solution due to the imprecise gradients (see Tab.~\ref{tab:dynamic_finetune_tal}, \ref{tab:dynamic_finetune_videomae}, and \ref{tab:dynamic_finetune_pointcloud} in Sec.~\ref{sec:ablation}). Therefore, we need a \textit{dynamic finetuning} mechanism to adjust the coefficient values to a low-error point during finetuning.

\begin{table*}[t]
\centering
\caption{\textbf{Downstream tasks experimented in this paper.} These tasks involve various pretraining methods, including fully-supervised  classification and self-supervised learning methods such as MAE~\cite{he2022masked}. They also use different backbones, e.g., Video Swin~\cite{vswin}, ViT~\cite{dosovitskiy2021image}. }
\vspace{-8pt}
\small
\begin{tabular}{l|lllll}
\toprule
 Data type & Pretraining  & Backbone & Downstream task & Downstream method & Downstream dataset         \\
\hline
\multirow{3}{*}{Video} & Classification & Video Swin - tiny~\cite{vswin} & Temporal action detection & VSGN~\cite{zhao2021video} & ActivityNet-v1.3~\cite{caba2015activitynet} \\
& Classification & Video Swin - tiny~\cite{vswin} & Refer. video object segment. & MTTR~\cite{botach2022end} & A2D-Sentences~\cite{gavrilyuk2018actor} \\
& VideoMAE~\cite{tong2022videomae} & Video ViT - small~\cite{tong2022videomae} & Action recognition & VideoMAE~\cite{tong2022videomae} & SthSth-v2~\cite{goyal2017something} \\
\hline
Point cloud & MAE~\cite{he2022masked} & ViT - small~\cite{dosovitskiy2021image} & Point cloud segmentation & Pix4Point~\cite{Pix4Point} & S3DIS~\cite{armeni20163d} \\
\hline
Image & Classification & Swin - tiny~\cite{liu2021swin} & Object detection & DINO~\cite{zhang2023dino} & MS-COCO~\cite{lin2014microsoft}\\
\bottomrule
\end{tabular}
\label{tab:tasks}
\end{table*}

\subsubsection{Dynamic finetuning} \label{sec:dynamic_ft}

As mentioned above, we identify two key factors that impact the effectiveness of \methodname{} finetuning:  (1)  the model's proximity to the pretrained network $\mathcal{M}_n$, and (2) gradient precision of the customized back-propagation. The former  needs a high $\alpha$ value and a low $\beta$ value (the top-right corner in Fig.~\ref{fig:grad_error}), whereas the later  needs a low $\alpha$ value and a high $\beta$ value (the bottom-left corner in Fig.~\ref{fig:grad_error}). This presents an apparent contradiction in the finetuning requirements.

Upon further examination, we find that the relative importance of these factors changes over the course of finetuning. Initially, Factor (1) is critical as \methodname{} begins finetuning with the pretrained parameters $\theta_n$. However, its significance diminishes as the network adapts over successive finetuning epochs. Conversely, Factor (2) becomes crucial in the later stages when seeking a precise solution, though it is less critical during the initial, more chaotic phases.

\textbf{First},  at the beginning of  finetuning, $\lambda \rightarrow 0, \alpha\rightarrow1$, i.e., the architecture needs to be as close as possible to the original pretraining network $\mathcal{F}$,  to have a matched initialization. But this becomes less important at later finetuning stages when the network has already evolved over iterations of training. \textbf{Second}, at later stages  when the optimal solution is being sought, $\lambda \rightarrow 1, \alpha\rightarrow0$, i.e., the architecture needs to be as close as possible to the reversible network, to ensure precise back propagation. But this doesn't matter as much at early stages  when the loss itself is high. 

Based on these analyses, we propose a dynamic finetuning mechanism for \methodname{}. At the beginning of finetuning, we use $\alpha=1, \beta=0.1$, and initialize \methodname{} with the parameters $\theta_n$ of the pretrained model $\mathcal{M}_n$. Then during the finetuning process, we progressively decrease the value of $\alpha$ and increase the value of $\beta$ until an acceptable gradient error level is reached, a point we call an update end point, \eg $10^{-7}$ as shown in Fig.~\ref{fig:grad_error}.  
After that, we use the fixed $\alpha$ and $\beta$ values for the rest of the training iterations. The next question is how to schedule the decrease of $\alpha$ and the increase of $\beta$ to have the lowest impact to the accuracy.

\textbf{Updating schedule: $\alpha$ first or $\beta$ first?} There are three options of updating the values of $\alpha$ and $\beta$: (1) fully updating $\alpha$ before updating $\beta$, (2) fully updating $\beta$ before updating $\alpha$, and (3) simultaneously updating $\alpha$ and $\beta$. Seen from Fig.~\ref{fig:grad_error}, the
shortest path connecting the start point (the top-right corner) and the update end point is along the diagonal signified by the blue arrows. This path reaches the update end point, which has a desired gradient error level, at earlier iterations than the paths in vertical or horizontal directions. This diagonal path corresponds to the third option --- simultaneously updating  $\alpha$ and $\beta$. We will compare the performance of the three options in Sec.~\ref{sec:ablation}.

\textbf{Updating policy: what functions to compute the values of $\alpha$ and $\beta$?} We consider  $\alpha$ and $\beta$ as functions of training iterations,  $\alpha$ being a monotonically non-increasing function and $\beta$ being a monotonically non-decreasing function. $\alpha$ always starts from 1 and $\beta$ always from 0.1. They end changing at the update end point, \eg $\alpha=0.3, \beta=0.7$ as illustrated in Fig.~\ref{fig:grad_error}. During our experiments, we find that simple linear functions are effective (see Sec.~\ref{sec:ablation}), and therefore employ linear functions for all the tasks.

\textbf{Updating frequency and end epoch.} Under our updating policy, there are two hyper-parameters. The first one is the updating frequency $\eta$, which means that the two coefficients are updated every $\eta$ epochs / iterations. The second one is the end epoch $\tau$, which means that the two coefficients reach the update end point at the $\tau^{th}$ epoch. We provide the comparison of different $\eta$ values for the task of action recognition with VideoMAE pretrained model~\cite{tong2022videomae} in Sec.~\ref{sec:ablation}, and have the choices of the two hyper-parameters for all tasks in the appendix.

\begin{table*}[t]
\centering
\caption{\textbf{Memory and accuracy comparison on different video understanding tasks.} Conventional: conventional non-reversible backbone; Reversible: previous reversible backbone~\cite{zhao2023re2tal}; Hard finetune: directly initializing the reversible backbone using pretrained parameters. ``mAP": mean average precision,  ``Acc": accuracy. } 
\vspace{-8pt}
\setlength{\tabcolsep}{1.8pt}
\small
\begin{tabular}{ll|ca|ca|ca}
\toprule
\multirow{2}{*}{Downstream training} && \multicolumn{2}{c|}{Temporal action detection~\cite{zhao2021video}}               & \multicolumn{2}{c|}{Refer. video object segmentation~\cite{botach2022end}}               & \multicolumn{2}{c}{Action recognition~\cite{tong2022videomae}}        \\
  &  & \multicolumn{1}{c}{Avg. mAP} & \multicolumn{1}{a|}{Memory} & \multicolumn{1}{c}{mAP} & \multicolumn{1}{a|}{Memory} & \multicolumn{1}{c}{Top-5 acc} & Memory  \\
\hline
\multirow{1}{*}{Conventional} & Frozen backbone  & 34.4\%   & 12.2GB   & 41.2\%   & 9.1GB   & 29.9\% &   2.4GB       \\
& End-to-end    & 36.2\%   & 44.7GB  & 44.1\%     & 41.5GB    & 90.3\%  &  29.3GB     \\
\hline
\multirow{1}{*}{Reversible~\cite{zhao2023re2tal}}    &From scratch & 28.3\%     & 24.1GB  & /    & 18.0GB   & 56.8\%    &   6.0GB     \\
& Hard finetune   & 35.4\%        & 24.1GB        & 42.1\%    & 18.0GB       & 78.9\%        &  6.0GB     \\
\textbf{Dr$^2$Net}    & \textbf{End-to-end }  & \textbf{36.3}\%                          & 24.1GB                          & \textbf{42.9}\%                          & 18.0GB     & \textbf{89.0}\%   &  6.0GB  \\
\bottomrule
\end{tabular}\label{tab:mem_acc_comp_video}\\
\footnotesize{
$^*$ We reproduced the results of conventional end-to-end finetuning, using the official code of each method for fair comparison.}
\end{table*}

\begin{table}[t]
\caption{\textbf{Memory and accuracy comparison on point cloud segmentation and object detection.}  Convent.: conventional non-reversible backbone; Reversible: previous reversible backbone~\cite{zhao2023re2tal}; Hard finetune: directly initializing the reversible backbone using pretrained parameters. “mIoU”: mean intersection over union,  “mAP”: mean average precision. Note that \textit{frozen backbone} for \textit{conventional} doesn't save much memory for point cloud segmentation since the downstream method Pix4Point~\cite{Pix4Point} has added parameters before the network backbone.} 
\vspace{-8pt}
\setlength{\tabcolsep}{2.5pt}
\small
\centering
\begin{tabular}{ll|ca|ca}
\toprule
\multicolumn{2}{l|}{Downstream training}  & \multicolumn{2}{l|}{ Point cloud seg.~\cite{Pix4Point} }      & \multicolumn{2}{l}{Object dect.~\cite{zhang2023dino}}                        \\
& &\multicolumn{1}{c}{mIoU} & \multicolumn{1}{c|}{Memory} & \multicolumn{1}{c}{mAP} & \multicolumn{1}{c}{Memory } \\
\hline
Convent. & Frozen backb.  & 62.0\%    & 21.2GB  & 49.7\%  & 26.9GB                        \\ 
& End-to-end      & 69.6\%    & 22.5GB   & 51.3\%         & 54.0GB                  \\
\hline
Reversible    & From scratch & 65.7\%   & 15.6GB                         & 38.7\%  & 30.0GB                           \\
& Hard finetune        & 62.5\% & 15.6GB & 49.6\%    & 30.0GB                           \\
\textbf{Dr$^2$Net}  & End-to-end    & \textbf{68.1}\%        & 15.6GB    & \textbf{51.3}\%                         & 30.0GB    \\
\bottomrule
\end{tabular}\label{tab:mem_acc_comp_other}\\
\footnotesize{
$^*$ We reproduced the results of conventional end-to-end finetuning, using the official code of each method for fair comparison.}
\end{table}

\section{Experiments}
\label{sec:exp}

We conducted extensive experiments on various tasks to show the effectiveness of our proposed Dr$^2$Net, and summarize them in Tab.~\ref{tab:tasks}.  We target 5 different kinds of vision tasks that require high-dimensional data such as videos, or high-resolutional images such as point cloud. As listed in the table, these tasks use different downstream datasets, and adopt different backbones, including Swin~\cite{liu2021swin}, ViT~\cite{dosovitskiy2021image}, Video Swin~\cite{vswin} and Video ViT~\cite{tong2022videomae}, which have been pretrained in different ways, such as fully supervised classification and self-supervised learning  MAE~\cite{he2022masked}. We provide the implementation details of these tasks in the appendix.

\subsection{Effectiveness of Dr$^2$Net} \label{sec:exp_effect}

In Tab.~\ref{tab:mem_acc_comp_video} and Tab.~\ref{tab:mem_acc_comp_other}, we show the effectiveness of our Dr$^2$Net in memory saving as well as in accuracy preservation, by comparing to conventional finetuning and other reversible methods on all the observed tasks listed in Tab.~\ref{tab:tasks}. Conventional finetuning uses the same non-reversible network as pretraining, and therefore consumes a large amount of GPU memory with end-to-end finetuning (Row 2). If we freeze the backbone and only train the downstream task-specific layers (Row 1), memory usage is dramatically reduced, but at the same time accuracy is also significantly impaired. Previous reversible models (e.g., \cite{zhao2023re2tal}) cannot directly finetune from pretrained non-reversible model, and have to train from scratch (Row 3). Re$^2$TAL~\cite{zhao2023re2tal} supports reusing the architecture of the pretrained model, and with it we tried \textit{hard finetune} by initializing the rewired reversible model using the pretrained parameters (Row 4). We can see that training the reversible model from scratch on the downstream tasks lead to obviously inferior performance. With \textit{hard finetune}, better performance is achieved, but there is still a big gap from conventional end-to-end finetuning. 

Seen from Tab.~\ref{tab:mem_acc_comp_video} and Tab.~\ref{tab:mem_acc_comp_other}, our proposed Dr$^2$Net (Row 5), saves 46.1\%, 56.6\%, and 79.5\% memory for the three video tasks respectively, and saves 30.6\% and 44.4\% memory for point cloud segmentation and object detection respectively. Across these experiments, our Dr$^2$Net reaches comparable accuracy to the original network while significantly reducing memory consumption when finetuning end-to-end. In these experiments, we use the smallest network variants for each type of backbone due to limited computational resources. Note that the memory saving will be more significant with deeper networks since reversible networks use constant memory regardless of network depths~\cite{revvit,zhao2023re2tal}.

Theoretically, the reversible training adds about 33\% more operations, as pointed out in RevNet~\cite{revnet}, but the actual latency can be smaller, varying among tasks. Table~\ref{tab:time} compares the training time of 
conventional end-to-end training and Dr$^2$Net for all the five tasks. 

\begin{table}[t]
\centering
\caption{\textbf{Training time comparison of Dr$^2$Net to conventional end-to-end finetuning. } The numbers are training time per epoch.}
\vspace{-3mm}
\setlength{\tabcolsep}{4.5pt}
\footnotesize
\begin{tabular}{l|ccccc} 
 \hline
Task & TAD~\cite{zhao2021video} &RVOS~\cite{botach2022end}&AR~\cite{tong2022videomae} &PCS~\cite{Pix4Point} &OD~\cite{zhang2023dino}\\
\hline
Conventional & 198 min &24 min &50 min  &178 sec &206 min \\
{Dr$^2$Net} &{261 min} &{29 min} &{89 min}& {186 sec} &{216 min}  \\ 
 \hline
\end{tabular}\label{tab:time}
\vspace{-5mm}
\end{table}

\subsection{Ablation Study and Design Analysis}\label{sec:ablation}

We perform the following ablation study and design analysis on multiple tasks to validate our design choices.

\textbf{Dynamic finetuning} ablation is shown in Tab.~\ref{tab:dynamic_finetune_tal}, Tab.~\ref{tab:dynamic_finetune_videomae}, and Tab.~\ref{tab:dynamic_finetune_pointcloud} for the tasks of temporal action detection~\cite{zhao2021video}, action recognition~\cite{tong2022videomae}, and point cloud segmentation~\cite{Pix4Point}, respectively. Our \methodname{} uses dynamic finetuning that updates the values of the two coefficients $\alpha$ and $\beta$ during the finetuning process, as described in Sec.~\ref{sec:fintuning}. We compare it to using vanilla finetuning, which uses fixed values of  $\alpha=1$ and $\beta=0.1$ throughout the finetuning process. From the tables, we can see that Dynamic finetuning leads to obviously higher accuracy than vanilla finetuning for all the tasks. The advantage of dynamic finetuning is significantly evident with the VideoMAE~\cite{tong2022videomae} pretrained model, as shown in Tab.~\ref{tab:dynamic_finetune_videomae}. Without dynamic finetuning,  action recognition using the VideoMAE pretrained model (Row 1 in Tab.~\ref{tab:dynamic_finetune_videomae}) is even worse than training from scratch (Row 3 in Tab.~\ref{tab:mem_acc_comp_video}).

\textbf{$\alpha$ and $\beta$ updating schedules} are compared in Tab.~\ref{tab:update_schedule_videomae}, Tab.~\ref{tab:update_schedule_objectdetection}, and Tab.~\ref{tab:update_schedule_pointcloud} for the tasks of action recognition~\cite{tong2022videomae}, object detection~\cite{zhang2023dino}, and point cloud segmentation~\cite{Pix4Point}, respectively. Our Dr$^2$Net updates $\alpha$ and $\beta$ simultaneously instead of finishing updating one  before updating the other, as described in Sec.~\ref{sec:fintuning}. We compare our simultaneous updating schedule to the following two schedules with the same updating frequency: (1) update $\alpha$ first until it reaches the updating end point, and then update $\beta$ (Row 1); (2) update $\beta$ first until it reaches the updating end point, and then update $\alpha$ (Row 2). Empirically, we find that the gradient error level of $10^{-7}$ (\ref{fig:grad_error}) can preserve the accuracy very well, therefore we consider the  $\alpha$ and  $\beta$ values at this point as the update end point. From the three tables, we can tell that our simultaneous updating schedule gives higher performance than the other schedules.

\begin{table}
\centering
\caption{\textbf{Ablation study of dynamic finetuning on temporal action detection~\cite{zhao2023re2tal}.} Vanilla finetuning uses fixed values of  $\alpha=1$ and $\beta=0.1$ throughout finetuning, our \methodname{} uses dynamic finetuning, which updates the two coefficients dynamically.} 
\vspace{-8pt}
\setlength{\tabcolsep}{1.3pt}
\small
\begin{tabular}{l|ccca} 
\toprule
Method &  0.5 & 0.75  & 0.95  & Avg. mAP\\
\hline  
{Vanilla finetune } &52.25\% &35.86\%&10.01\%&35.44\%\\ 
{\textbf{Dynamic finetune (Dr$^2$Net)}} &\textbf{53.24\%} & \textbf{36.97\%}& \textbf{10.16\%}&\textbf{36.27\%}\\
\bottomrule
\end{tabular}
\label{tab:dynamic_finetune_tal}
\vspace{-2pt}
\end{table}

\begin{table}
\centering
\caption{\textbf{Ablation study of dynamic finetuning on action recognition with VideoMAE pretrained model~\cite{tong2022videomae}.} Vanilla finetuning  uses fixed values of  $\alpha=1$ and $\beta=0.1$ throughout the finetuning, our \methodname{} uses dynamic finetuning, which updates the two coefficients dynamically.} 
\vspace{-8pt}
\small
\begin{tabular}{l|cc} 
\toprule
Method &  Top-1 Acc  & Top-5 Acc  \\
\hline  
{Vanilla finetune } & 26.48\% & 53.08\%\\ 
{\textbf{Dynamic finetune (Dr$^2$Net)}} &\textbf{64.57\%} & \textbf{89.01\%}\\
\bottomrule
\end{tabular}
\label{tab:dynamic_finetune_videomae}
\end{table}

\begin{table}
\centering
\caption{\textbf{Ablation study of dynamic finetuning on point cloud segmentation~\cite{Pix4Point}.} Vanilla finetuning  uses fixed values of  $\alpha=1$ and $\beta=0.1$ throughout the finetuning, our \methodname{} uses dynamic finetuning, which updates the two coefficients dynamically. }
\vspace{-8pt}
\small
\begin{tabular}{l|cc} 
\toprule
Method &   mIoU \\
\hline  
{Vanilla finetune } &  57.57\%\\ 
{ \textbf{Dynamic finetune (Dr$^2$Net)}} &\textbf{ 68.13\%} \\
\bottomrule
\end{tabular}
\label{tab:dynamic_finetune_pointcloud}
\end{table}

\begin{table}
\centering
\caption{\textbf{Comparison of $\alpha$ or $\beta$ updating schedules on action recognition with the VideoMAE pretrained model~\cite{tong2022videomae}.} ``Acc" means accuracy. Simultaneously updating  $\alpha$ and $\beta$ leads to the highest accuracy. } 
\vspace{-8pt}
\small
\begin{tabular}{l|cc} 
\toprule
Schedule &  Top-1 Acc & Top-5 Acc\\
\hline
{$\alpha$ first, $\beta$ second } & 60.51\% & 86.64\%\\
{$\beta$ first, $\alpha$ second} & 58.40\% & 85.30\%\\ 
\textbf{$\alpha$ and $\beta$ simultaneously (Dr$^2$Net)} & \textbf{64.57\%} & \textbf{89.01\%} \\
\bottomrule
\end{tabular}
\label{tab:update_schedule_videomae}
\end{table}

\begin{table}
\centering
\caption{\textbf{Comparison of $\alpha$ or $\beta$ updating schedules on object detection~\cite{zhang2023dino}.} Simultaneously updating  $\alpha$ and $\beta$ leads to the highest accuracy.} 
\vspace{-8pt}
\small
\begin{tabular}{l|cc} 
\toprule
Schedule & mAP\\
\hline
{$\alpha$ first, $\beta$ second } & 50.2\%\\
{$\beta$ first, $\alpha$ second} & 50.7\% \\ 
\textbf{$\alpha$ and $\beta$ simultaneously (Dr$^2$Net)} & \textbf{51.3\%} \\
\bottomrule
\end{tabular}
\label{tab:update_schedule_objectdetection}
\end{table}

\begin{table}
\centering
\caption{\textbf{Comparison of $\alpha$ and  $\beta$ updating schedules on 3D point cloud segmentation~\cite{Pix4Point}.} Simultaneously updating  $\alpha$ and $\beta$ leads to the highest accuracy.} 
\vspace{-8pt}
\small
\begin{tabular}{l|c} 
\toprule
Schedule &  mIoU \\
\hline
{$\alpha$ first, $\beta$ second } & 66.40\%\\
{$\beta$ first, $\alpha$ second} & 64.90\%\\ 
\textbf{$\alpha$ and $\beta$ simultaneously (Dr$^2$Net)} & \textbf{68.13\%} \\
\bottomrule
\end{tabular}
\label{tab:update_schedule_pointcloud}
\vspace{-10pt}
\end{table}

\textbf{Updating frequency $\eta$}  is studied in Tab.~\ref{tab:freq_videomae} for the task of action recognition~\cite{tong2022videomae}. The two coefficients $\alpha$ and $\beta$ are updated every $\eta$ iterations. A smaller value of  $\eta$ means that they are updated more frequently and in a smaller step. We can see from the table that the smaller $\eta$ is, the higher performance we will obtain.

\textbf{Updating policies} are compared in Tab.~\ref{tab:policy} for the task of temporal action detection~\cite{zhao2021video}. If we consider the values of $\alpha$ and $\beta$ to be functions of training iterations, we can use different functions to represent different updating policies. We compare the linear updating policy in our \methodname{} to the exponential and logarithm policy in the table, and find that our linear policy leads to the best performance.

\begin{table}[t]
\centering
\caption{\textbf{Comparison of different values of the updating frequency $\eta$ on action recognition with VideoMAE pretrained model~\cite{tong2022videomae}}. The two coefficients $\alpha$ and $\beta$ are updated every $\eta$ iterations. Smaller $\eta$ values means more frequent update, and it shows higher accuracy than larger $\eta$ values.}
\vspace{-8pt}
\small
\begin{tabular}{l|ccccc} 
\toprule
$\eta$ &  2 iter & 5 iter & 20 iter & 50 iter & 100 iter  \\
\hline
{Top-5 Acc (\%)} & \textbf{89.01} & 88.75 &  88.61 & 88.49 & 88.08\\
\bottomrule
\end{tabular}
\label{tab:freq_videomae}
\end{table}

\begin{table}
\centering
\caption{\textbf{Comparison of different updating policies for $\alpha$ and $\beta$.} We consider the values of $\alpha$ and $\beta$ as functions of training iterations, and experiment with the following three functions as different updating policies. We report mean average precision (mAP) at tIoU thresholds 0.5, 0.75, and 0.95, and average mAP. The linear function shows the highest accuracy.} 
\vspace{-8pt}
\setlength{\tabcolsep}{1.5pt}
\small
\setlength{\tabcolsep}{4.5pt}
\begin{tabular}{l|ccca} 
\toprule
Policy &  0.5 & 0.75  & 0.95  & Avg. mAP\\
\hline
{Exponential} &52.68\% &35.93\%&9.47\%& 35.46\%\\
{Logarithm} &52.43\%&36.49\%&9.38\%& 35.69\%\\ 
{\textbf{Linear (Dr$^2$Net)}} &\textbf{53.24\%} & \textbf{36.97\%}  &\textbf{10.16\%} &\textbf{36.27\%}\\
\bottomrule
\end{tabular}
\label{tab:policy}
\vspace{-10pt}
\end{table}

\section{Conclusions}
\label{sec:conclude}

In this paper, we propose Dynamic Reversible Dual-Residual Networks (\methodname{}), a novel approach for fine-tuning pretrained  models with significantly reduced memory usage.
\methodname{} contains two types of residual connections, one maintaining the residual structure in the pretrained models, and the other introducing reversibility to enable  clearing of intermediate activations from memory during training. 
We adopt a dynamic finetuning strategy that ensures a smooth transition from the non-reversible pretrained network to the reversible network. 
Evaluation across various tasks demonstrates that \methodname{} achieves performance comparable to conventional finetuning methods but with much lower memory requirements. 

This work presents a practical solution for scenarios where downstream tasks are hindered by excessive memory consumption or restricted memory capacity. This includes applications involving large models, tasks dealing with high-resolution or high-dimensional data, and on-device learning environments. It could open avenues for future research in memory-efficient network architectures within the field of computer vision, as well as extending its implications to applications beyond computer vision, including natural language processing and audio analysis. 

\vspace{10pt}

\noindent \textbf{Acknowledgement.} This work was supported by the King Abdullah University of Science and Technology (KAUST) Office of Sponsored Research through the Visual Computing Center (VCC) funding, as well as the SDAIA-KAUST Center of Excellence in Data Science and Artificial Intelligence (SDAIA-KAUST AI).

\clearpage
\appendix

\begin{center}
\centering\textbf{\LARGE Appendix} 
\end{center}

\vspace{0.2cm}

\renewcommand\thesection{\Alph{section}}
\renewcommand\thesubsection{\thesection.\arabic{subsection}}
\setcounter{section}{0}

In the paper, we have described the core techniques of Dr$^2$Net, and 
provided the key experiments that support our contributions. In this appendix, we provide additional details of the method and the experiment implementation, as well as extra experimental results.

\section{Additional Details of the Method}
\subsection{Proof of invertibility of Dr$^2$Net}

Our proposed Dr$^2$Net, as illustrated in  Fig.~\ref{fig:unify_res_rev} and Eq.~\ref{eq:forward_lambda}, is a reversible network, and mathematically, an invertible function. In this section, we mathematically prove its invertibility. Let's rewrite the computation of the $i^{th}$ module (Eq.~\ref{eq:forward_lambda}) in the following equation for clarity.
\begin{equation}
\left\{\begin{array} { l }
{ y _ { i } = \beta \times x _ { i-1 }  } \\
{ x _ { i } = \mathcal{G }_i ( x_{ i-1 } ) + y_{ i-1 } }.
\end{array}  \right.
\label{eq:forward_lambda_sup}
\end{equation}
Let's make $I = (x_{ i-1 }, y_{ i-1 })$, which represents the input activations to the $i^{th}$ module, and make $O = (y_{ i }, x_{ i })$, which represents the output activations from the $i^{th}$ module. The Jacobian matrix of Eq.~\ref{eq:forward_lambda_sup} is computed as follows
\begin{equation}
J = \frac{\partial O}{\partial I} = 
\begin{bmatrix}
\frac{\partial y_i}{\partial x_{i-1}} & \frac{\partial y_i}{\partial y_{i-1}} \\
\\
\frac{\partial x_i}{\partial x_{i-1}} & \frac{\partial x_i}{\partial y_{i-1}} 
\end{bmatrix} 
=
\begin{bmatrix}
\beta\times I_d & 0 \\
\\
\frac{\partial G_i}{\partial x_{i-1}} & I_d
\end{bmatrix}.
\label{eq:jacobian}
\end{equation}
In Eq.~\ref{eq:jacobian}, $I_d$ is the identity matrix of size $d$, where $d$ is the dimension of the activations $x_i, y_i, x_{i-1}, y_{i-1}$. Its determinant is computed as 
\begin{equation}
\det(J)=\det(\beta\times I_d) \cdot \det(I_d) =\beta^d.
\label{eq:det}
\end{equation}
As described in the paper, $\beta \neq 0$, and hence, the Jacobian determinant
$\det(J)$ is not zero. Therefore, the function in Eq.~\ref{eq:forward_lambda_sup} representing the $i^{th}$ module in Dr$^2$Net is invertible. 

If we stack multiple such reversible modules, represented by the above invertible functions, without inserting any downsampling operations, we will form \textit{a stage} in \methodname{}. 
One stage is mathematically composition of such invertible functions, and therefore, the entire stage of Dr$^2$Net is also invertible. Between stages where there are downsampling operations, we cache the activations after each stage following \cite{revvit,zhao2023re2tal}.

\subsection{Illustration of the reverse computation}

\begin{figure*}[t]
\begin{center}
\footnotesize
\includegraphics[width=0.75\textwidth]{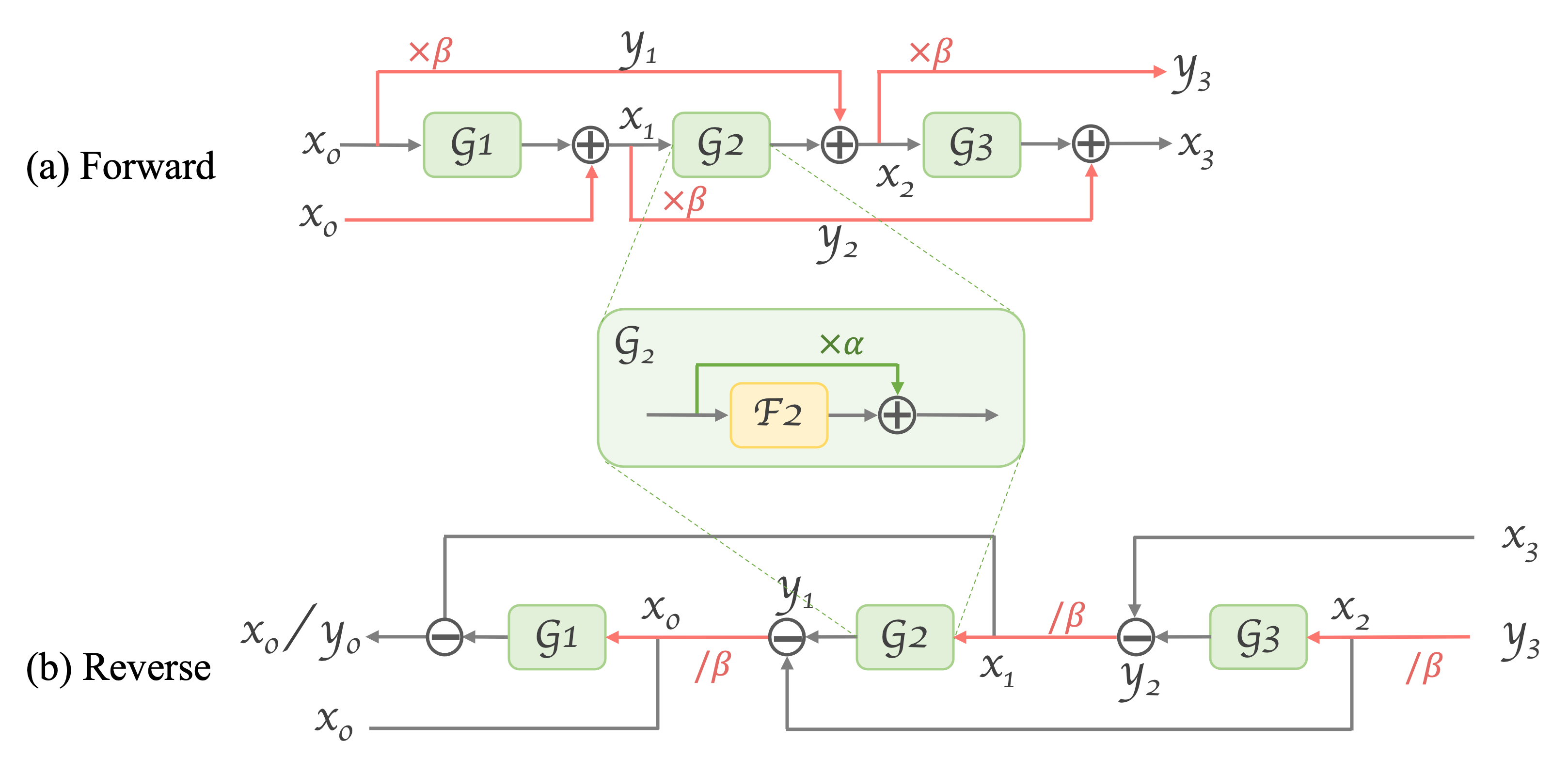}
\end{center}
\vspace{-15pt}
\caption{\textbf{Forward and reverse computation in Dr$^2$Net.} Gray arrows denote the pathway for $x_i$, and pink arrows denote the pathway for $y_i$. Compared to Fig.~\ref{fig:unify_res_rev}, we place the $\mathcal{F}_i$ blocks along with their $\alpha$-weighted residual connections inside the module $\mathcal{G}_i$.}
\label{fig:reverse}
\begin{align}
\textrm{Forward:} \quad
\left\{\begin{array} { l }
{ y _ { 1 } = \beta \times x _ { 0 }  } \\
{ x _ { 1 } = \mathcal{G}_1 ( x_{ 0 } ) + y_{ 0 } },
\end{array} 
\Rightarrow 
\left\{\begin{array}{l}
y_2=\beta \times x_1 \\
x_2=\mathcal{G}_2(x_1)+y_1,
\end{array}\right. 
\Rightarrow 
\left\{\begin{array}{l}
y_3=\beta \times x_2 \\
x_3=\mathcal{G}_3(x_2)+y_2.
\end{array}\right.\right.
\end{align}
\label{eq:forward_expand_sup}
\end{figure*}

In  Fig.~\ref{fig:unify_res_rev} (c), we have illustrated the architecture of our Dr$^2$Net with the $\mathcal{F}$ blocks and the two types of residual connections. In Fig.~\ref{fig:reverse} (a), we re-illustrate this forward process by moving the  $\mathcal{F}$ blocks along with their $\alpha$-weighted residual connection inside the $G$ blocks for conciseness and to be consistent with Eq.~\ref{eq:forward_lambda}. In Fig.~\ref{fig:reverse} (b), we illustrate its corresponding reverse process. 

\begin{figure}[t]
\begin{center}
\footnotesize
\includegraphics[width=0.49\textwidth]{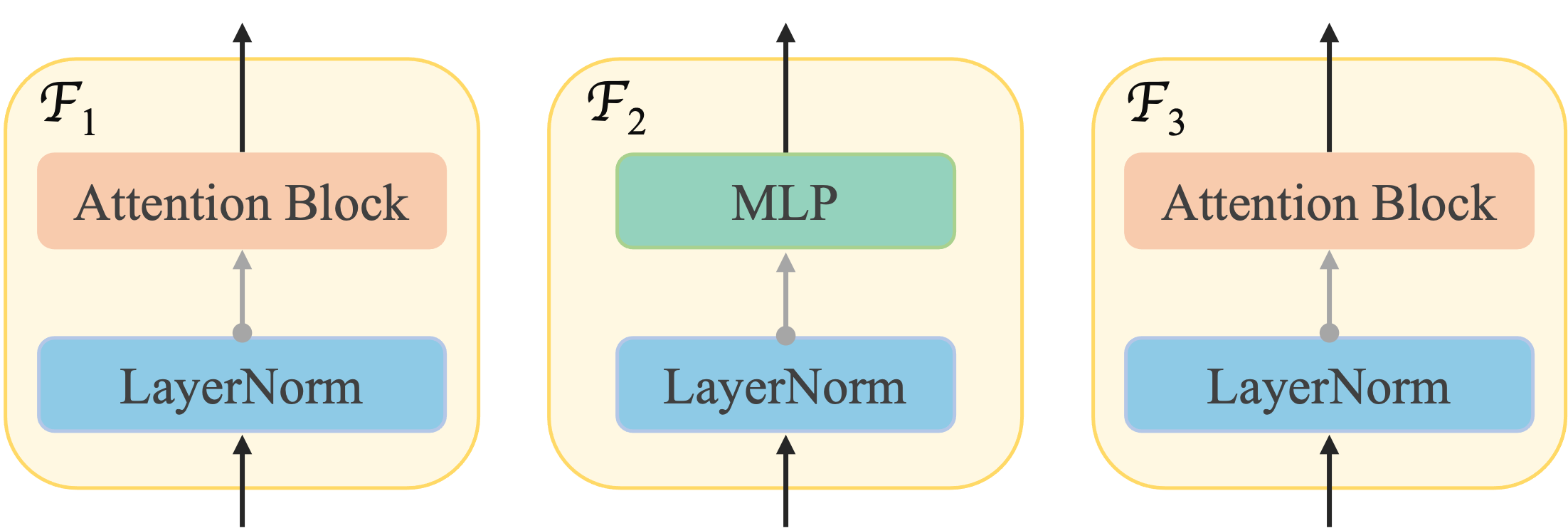}
\end{center}
\vspace{-15pt}
\caption{\textbf{$\mathcal{F}_i$ blocks in a transformer network.} If the pretrained model is a transformer network, e.g., Swin~\cite{liu2021swin} or ViT~\cite{dosovitskiy2021image}, the $\mathcal{F}_i$ blocks in our Dr$^2$Net are attention layers or MLP layers. The two types of layers are interleaved, namely, if $\mathcal{F}_1$ is an attention layer, then $\mathcal{F}_2$ is an MLP layer, and $\mathcal{F}_3$ is an attention layer, and so on.}
\label{fig:F_block}
\end{figure}

For detailed mathematical formulation of the forward and reverse processes,  we  expand  Eq.~\ref{eq:forward_lambda} as Eq.~{\color{red}6},  and Eq.~\ref{eq:reverse_lambda} as  Eq.~{\color{red}7}  to illustrate the computation in three modules. In the equations, $\mathcal{G}_i(x_{i-1})=\mathcal{F}_i(x_{i-1})+\alpha\times x_{i-1}$.

We can see from Fig.~\ref{fig:reverse} (b) and Eq.~{\color{red}7} that during the reverse computation, given $x_i$ and $y_i$ where $i=3$, we will compute all the intermediate activations $x_i, y_i$ where $i=0,1,2$ module by module. In the $i^{th}$ module, $x_{i-1}$ is computed first using $x_{i-1} = y_{i} / \beta$. Then $x_{i-1}$ is used to compute $\mathcal{G}_{i}(x_{i-1})$ to finally compute $y_{i-1}$.

\begin{figure*}[t]
\begin{align}
\textrm{Reverse:} \quad
\left\{\begin{array}{l}
x_0= y_1 / \beta \\
y_0=x_1-\mathcal{G}_1(x_0),
\end{array}
\Leftarrow
\left\{\begin{array} { l }
{ x _ { 1 } = y _ { 2 } / \beta} \\
{ y _ { 1 } = x _ { 2 } - \mathcal { G }_2 ( x_ { 1 } ) },
\end{array}\right.
\Leftarrow
\left\{\begin{array} { l }
{ x _ { 2 } = y _ { 3 } / \beta} \\
{ y _ { 2 } = x _ { 3 } - \mathcal { G }_3 ( x_ { 2 } ) }.
\end{array}\right.
\right.
\end{align} \label{eq:reverse_expand_sup}
\vspace{-15pt}
\end{figure*}

\subsection{Illustration of different types of ${F}$ blocks}

The basic blocks $\mathcal{F}_i$ in Dr$^2$Net, as illustrated in Fig.~\ref{fig:reverse}, can be any network block that doesn't change the feature dimensions. We use $\mathcal{F}_i$ and $\mathcal{F}$ interchangeably in the following text. The $\mathcal{F}_i$ blocks can be instantiated as different types of blocks when the pretrained networks have different architectures. In Fig.~\ref{fig:F_block}, we illustrate the $\mathcal{F}_i$ blocks of the popular transformer architectures, Swin~\cite{liu2021swin} and ViT~\cite{dosovitskiy2021image}. In this case, the $\mathcal{F}_i$ blocks in our Dr$^2$Net are attention layers or MLP layers. The two types of layers are interleaved, namely, if $\mathcal{F}_1$ is an attention layer, then $\mathcal{F}_2$ is an MLP layer, and $\mathcal{F}_3$ is an attention layer, and so on.

\subsection{Gradient errors of different networks}

\begin{figure*}[t]
\begin{center}
\footnotesize
\includegraphics[width=0.8\textwidth]{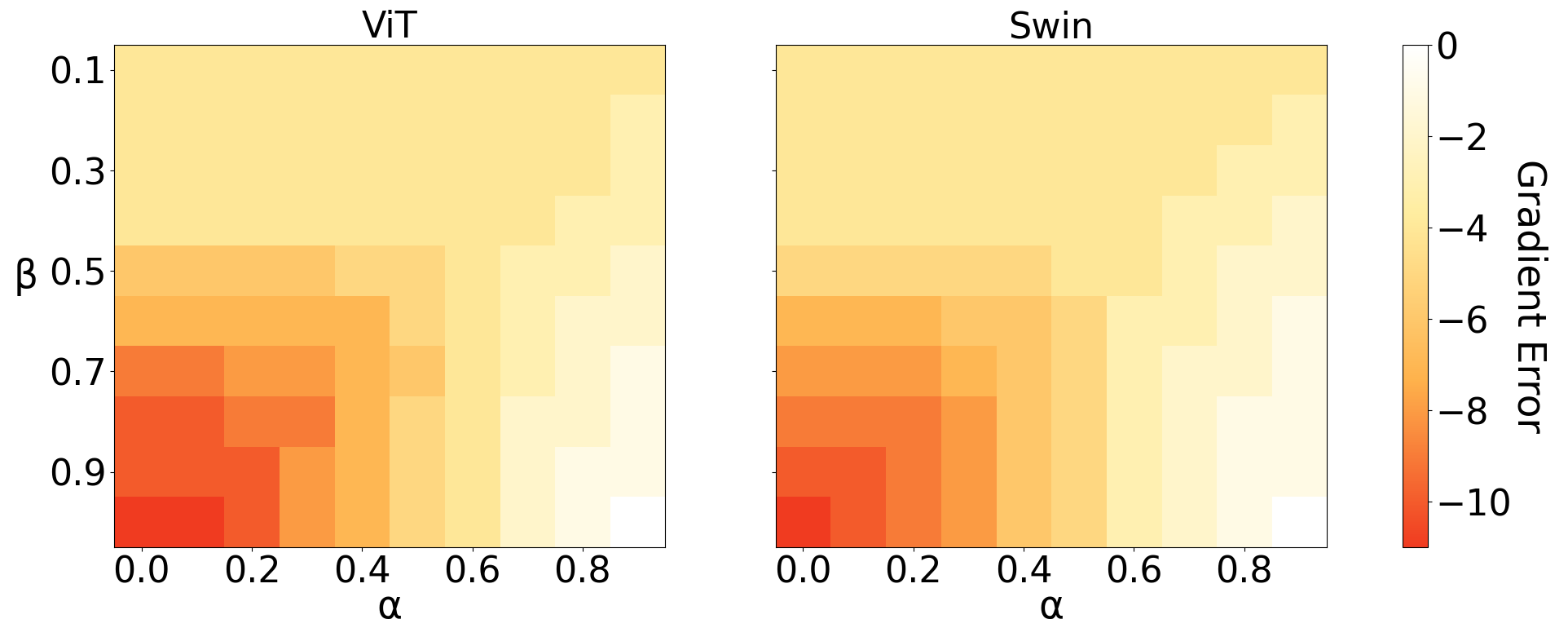}
\end{center}
\vspace{-15pt}
\caption{\textbf{Gradient error levels with different $\alpha$ and $\beta$ values for Video ViT-small and Video Swin-tiny.} The error levels of the two types of networks are similar, with the lowest in the bottom-left corners, and the highest in the bottom-right corners. Swin has slight lower error levels.}
\label{fig:error_vit_swin}
\end{figure*}

In the paper, we have illustrated the gradient error levels of video Swin-tiny~\cite{vswin} in Fig.~\ref{fig:grad_error}. In this subsection, we plot the error levels for another popular type of network Video ViT~\cite{tong2022videomae}, and provide more detailed explanations about the error maps.

In Fig.~\ref{fig:error_vit_swin}, we plot the error levels of the two types of networks Video ViT-small (used in VideoMAE~\cite{tong2022videomae}) and Video Swin-tiny, both with 12 layers. As we described in the paper,  customized back-propagation which computes gradients with recomputed intermediate activations through the reverse process (Eq.~\ref{eq:reverse_lambda}), is used to save memory for the reversible networks. 
This may introduce numerical errors that are accumulated due to floating point computation with limited precision. The idea of the gradient error levels  is to assess the precision of the customized back-propagation compared to using the default back-propagation that computes gradients with the activations cached in GPU memory. Concretely, the values in the gradient-error-level maps in Fig.~\ref{fig:error_vit_swin} are obtained as follows. Given one point $\alpha=\alpha_0$ and $\beta=\beta_0$, we obtain one Dr$^2$Net architecture that is adapted from Video ViT-small or Video Swin-tiny. For this \methodname{} architecture, we have two ways of implementations: (1) \methodname{}-A  with customized back-propagation, and (2) \methodname{}-B with default back-propagation. We generate a random tensor, and feed it into \methodname{}-A and \methodname{}-B separately, and compute two versions of gradients respectively: $G_A$ and $G_B$. We compare $G_A$ and $G_B$ using \textit{torch.allclose($G_A$, $G_B$, rtol=1e-05, atol=atol)}, and record the lowest \textit{atol} value that gives \textit{torch.allclose() == True} as the value  at $(\alpha=\alpha_0, \beta=\beta_0)$ in  the gradient-error-level maps.

As we see from Fig.~\ref{fig:error_vit_swin}, though Swin has slightly lower error levels than ViT, the error levels of the two types of networks
are quite close, with the lowest in the bottom-left corners, and the highest in the bottom-right corners. When we initialize  \methodname{} from the pretrained ViT or Swin, we set $\alpha=1, \beta=0.1$, meaning the finetuning starts from the top-right corners of the map, as we described in Sec.~\ref{sec:dynamic_ft} in the paper. Considering that the errors at the top-right corner are too high to effectively train the networks, i.e., $10^{-4}$ and $10^{-5}$ for ViT and Swin respectively, we need the dynamic finetuning strategy to adjust the values of $\alpha$ and $\beta$ to reach a point with sufficient precision, which is the bottom-left region. It can be observed from the maps that the shortest path to reach the bottom-left region with monotonically non-increased error levels is along the diagonal, meaning updating $\alpha$ and $\beta$ simultaneously. 

In addition, to make Dr$^2$Net with new values of $\alpha$ and $\beta$ benefit from Dr$^2$Net with previous values of $\alpha$ and $\beta$, we need to update the values of $\alpha$ and $\beta$ in small steps. We use $\eta$ to determine the updating frequency of both coefficients, as described in Sec.~\ref{sec:dynamic_ft} in the paper. Given the total number of epochs for which $\alpha$ and $\beta$ are updated,
a smaller $\eta$ value indicates the changes of $\alpha$ and $\beta$ are more frequent but more incremental each time. We have shown in Tab.~\ref{tab:freq_videomae} in the paper that a smaller $\eta$ value results in higher performance for the task of action recognition with the VideoMAE~\cite{tong2022videomae} pretrained model.

\section{Implementation Details}
In this section, we provide the implementation details of the downstream tasks we have experimented in the paper.

\subsection{Temporal action detection}
Temporal action detection (TAD)~\cite{lin2019bmn,xu2020g,zhao2023re2tal} is a typical long-form video understanding task, that needs to process a long sequence of video frames to identify all the action instances. Given a long video, the task of TAD outputs the category as well as the start and end timestamps of each action. A representative dataset for this task is the largescale dataset ActivityNet-v1.3~\cite{caba2015activitynet}, that uses mean Average Precision (mAP) at 10 tIoU thresholds  in the range [0.5, 0.95] as well as average mAP  as the evaluation metric.

In our experiment, we use a recent TAD method VSGN~\cite{zhao2021video} as the detector, and Video Swin-tiny pretrained with Kinetics-400 classification as the backbone. For all the experiments of this task in Tab.~\ref{tab:mem_acc_comp_video} in the paper, we use the same setup as follows. As network input, we use 512 input frames, evenly sampled from the entire video regardless of the original video duration. The frame resolution is $224\times224$. We use the augmentation following \cite{zhao2023re2tal}.
The backbone learning rate is $1e-5$, the detector learning rate is $1e-4$, and the batch size is 2. The total number of epochs is 20. For \methodname{}, the coefficient updating frequency is 3 epochs, and the updating ends at the 10$^{th}$ epoch.

\subsection{Video object segmentation}

Video object segmentation aims to separate the foreground objects from the background region of a video at the pixel level~\cite{bhat2020learning,cheng2022xmem}. Recently, referring video object segmentation (RVOS) has drawn more attention~\cite{liu2021cross,mcintosh2020visual,ning2020polar}. Given a sequence of video frames and a text query, ROVS aims 
to segment all objects in the video referred by the input text
prior to determining the referred instance~\cite{gavrilyuk2018actor}. In this paper, we evaluated our method on the dataset A2D-Sentence~\cite{gavrilyuk2018actor}, which contains 3,754 videos with 8 action classes.

In the experiments, we utilize the method MTTR~\cite{botach2022end} as the segmentation head and the Kinetics-400~\cite{carreira2017quo} pretrained Video Swin-tiny as the backbone. In MTTR, the window size is set to 10, and the total batch size is set to 6. The video frames are resized such that the short side is at least 320 pixels and the long side at most 576 pixels. The model is trained for 70 epochs. For \methodname{}, the coefficient updating frequency is set to 2 iterations, and the updating ends at the 10$^{th}$ epoch.

\subsection{Action recognition}
Action recognition~\cite{zisserman2017kinetics, goyal2017something, vswin, slowfast, tong2022videomae} is a fundamental task in video understanding, which aims to classify a video clip into an action category. Though it doesn't require as long input sequences as TAD, its input is still 3D video data and it uses spatio-temporal attention with Transformers, which consumes a large amount of GPU memory. Therefore, memory-efficient finetuning is  important. If we can save memory consumption during training, then we will be able to feed more input frames, use larger batch sizes, and train larger networks, which will lead to higher performance. 

For the experiments, we adopt the widely used large-scale video dataset Something-Something V2~\cite{goyal2017something}, which contains around 169k videos for training and 20k videos for
validation, with 174 motion-centric action classes. We report the top-1 and top-5 accuracies as the evaluation metrics.
We have two sets of experiments on the task of action recognition, Set-A with the Video ViT backbones pretrained with VideoMAE~\cite{tong2022videomae} (Sec.~\ref{sec:exp_effect} in the paper), and Set-B with Image ViT backbones pretrained with DINOv2~\cite{oquab2023dinov2} (Sec.~\ref{sec:dinov2}). Both sets of experiments use the dataset Something-Something V2~\cite{goyal2017something} and the finetuning recipe of VideoMAE~\cite{tong2022videomae} for the downstream finetuning. For both sets, the input video resolution is $224\times224\times16$, 
the batch size is 384, the learning rate $1e-3$, and the total number of epochs is 40. For \methodname{}, the coefficient updating frequency is 2 iterations, and the updating ends at the 5$^{th}$ epoch.

\subsection{Object detection}

Object detection involves identifying and locating potential objects within an image. A notable example of state-of-the-art object detection approaches is DINO~\cite{zhang2023dino}, 
which enhances the performance of the DETR-based framework by denoisng its anchor boxes. For the downstream task of object detection in our work, we use DINO as the detection head and employ Swin Transformer~\cite{liu2021swin} as the image backbone. We evaluate the model's performance using the mean Average Precision (mAP) metric on the COCO val2017 dataset~\cite{lin2014microsoft}.

In our experiments, we follow the training receipt of the original DINO. The Swin Transformer is pretrained on the ImageNet-22k dataset with the image classification task. We utilize 4 scales of feature maps to conduct the experiments. The short side of an input image is randomly resized between 480 and 800 pixels, and the long side is resized to at most 1333. The total batch size is 16, and the number of training epochs is 12. For \methodname{}, the  updating frequency of the two coefficients is 2 iterations, and the updating ends at the 5$^{th}$ epoch.

\subsection{3D point cloud segmentation}
3D point cloud segmentation is the process of classifying point clouds into multiple meaningful regions, where the points in the same region have the same label.  
We conduct extensive experiments in \emph{S3DIS} \cite{armeni20163d}, which is the mostly-used benchmark for large-scale point cloud segmentation. S3DIS consists of $6$ areas with 271 rooms, where area-5 is used in testing and the others are used in training. Each area is a large point cloud of a building. We used the same preprocessing as Pix4Point \cite{Pix4Point} to extract the point cloud per room, and leveraged sphere sampling to sample $16,384$ points as a batch in training and testing. 
Following the standard practice \cite{PointNeXt}, our model is optimized using the cross-entropy loss with label smoothing of $0.1$, the AdamW optimizer \cite{loshchilov2019adamw} with a learning rate 1e-4, a cosine learning rate scheduler, 10 warmup epochs,  weight decay 1e-5, the batch size $8$, and $600$ total training epochs.  We use data augmentation including rotation, scaling, color auto-contrast, and color dropping.  For \methodname{}, the coefficient updating frequency is 10 iterations, and the updating ends at the 50$^{th}$ epoch.

\section{Supplementary Experiments}

\subsection{More pretraining methods}\label{sec:dinov2}

In the paper, we have shown the effectiveness of our \methodname{} on models with different pretraining methods, including fully-supervised classification, self-supervised learning with MAE~\cite{he2022masked} and VideoMAE~\cite{tong2022videomae}. In this subsection, we demonstrate our results with one more pretraining method DINOv2~\cite{oquab2023dinov2}.

DINOv2 is a self-supervised learning method that pretrain an image model on a largescale image dataset. We use it for the downstream task action recognition on the dataset  Something-Something v2~\cite{goyal2017something}. Since the architecture of the DINOv2 model is  ViT~\cite{dosovitskiy2021image}, which is agnostic of input data dimensions, we can directly apply the same ViT architecture to the video data and compute spatio-temporal attention. Considering that the patch embedding layer was pretrained for images which are 2D data,  we inflate those convolutional kernels to 3D during initialization to perform tube embedding instead of patch embedding. In addition, we interpolate the position embedding to match the video dimension.  Our implementation of finetuning the DINOv2 model on  Something-Something v2  follows VideoMAE~\cite{tong2022videomae} for the setup of the spatio-temporal attention, tube embedding, and the training recipe.

We demonstrate the memory consumption and the recognition accuracy in Tab.~\ref{tab:mem_acc_dinov2}. Compared to conventional end-to-end finetuning (Row 2), our \methodname{} (Row 5) only uses less than 1/4 memory, and its accuracy surprisingly surpasses conventional finetuning by a large margin. Considering that the accuracies in the table are taken from the results of the 40$^{th}$ epoch following VideoMAE~\cite{tong2022videomae}, the training might not have fully converged. Still, that shows our \methodname{} at least converges faster. This might be due to the domain gap between the image pretraining and the video downstream task, and is worth further exploration.

\begin{table}[t]
\centering
\caption{\textbf{Memory and accuracy comparison on action recognition using DINOv2~\cite{oquab2023dinov2} pretrained models.} The backbone ViT-small is used. Conventional: conventional non-reversible backbone; Reversible: previous reversible backbone~\cite{zhao2023re2tal}; Hard: directly initializing the reversible network using pretrained parameters. } 
\setlength{\tabcolsep}{2.0pt}
\small
\begin{tabular}{ll|cca}
\toprule
\multicolumn{2}{l|}{Downstream training}& {Top-1 acc}&{Top-5 acc } & Mem (GB) \\
\hline
\multirow{1}{*}{Conventional} & Frozen$^*$ & 33.10\%&/ &/         \\
& End-to-end    &  55.18\% & 82.79\%  & 34.2\\
\hline
\multirow{1}{*}{Reversible~\cite{zhao2023re2tal}}    &Scratch    &  14.31\% & 33.96\%  & 8.0  \\ 
& Hard   &   37.29\% & 66.22\%  & 8.0\\
\textbf{Dr$^2$Net}    & \textbf{End-to-end }   &  \textbf{64.98\%} & \textbf{88.90\%} & 8.0\\
\bottomrule
\end{tabular}\label{tab:mem_acc_dinov2}\\
\footnotesize{
$^*$ Frozen: linear probing results from the DINOv2~\cite{oquab2023dinov2} paper.}
\end{table}

\subsection{Using larger networks}
Our \methodname{} can significantly reduce the GPU memory consumption during finetuning. Using the saved GPU memory, we can support a larger backbone network to reach higher accuracy. We experiment with larger backbones for the tasks of action recognition with DINOv2~\cite{oquab2023dinov2} pretrained models, action recognition with VideoMAE~\cite{tong2022videomae} pretrained models, and object detection with DINO~\cite{zhang2023dino}. We demonstrate the accuracy and the corresponding GPU memory consumption in Tab.~\ref{tab:acc_vs_memory_dinov2}, Tab.~\ref{tab:acc_vs_memory_videomae} and Tab.~\ref{tab:acc_vs_memory_dinov2_odetection}, respectively.

For the first two tasks (Tab.~\ref{tab:acc_vs_memory_dinov2} and Tab.~\ref{tab:acc_vs_memory_videomae}), which use ViT~\cite{dosovitskiy2021image} as the backbone, we apply \methodname{} to ViT-base in addition to ViT-small. Using the larger backbone ViT-base (Row 3), the accuracy is obviously increased for both tasks. Compared to both conventional finetuning (Row 1), \methodname{} uses still less than half  of the memory (16.6 GB \vs 34.2 GB, 13.0 GB \vs 29.3 GB), but reaches much higher performance.

For the task of object detection~\cite{zhang2023dino}, we apply \methodname{} to Video Swin-small and Video Swin-base in addition to Video Swin-tiny. Using the larger backbone Swin-small (Row 3), the accuracy is obviously increased, while the memory is almost the same (30.1 GB). Using a even larger backbone Swin-small, the accuracy is dramatically higher than conventional finetuning (54.7\% \vs 51.3\%), while memory cost is only $60\%$ of it (32.4 GB \vs 54.0 GB).

\begin{table}
\centering
\caption{\textbf{Accuracy versus memory for action recognition~\cite{goyal2017something} with DINOv2~\cite{oquab2023dinov2} pretrained models.} Our Dr$^2$Net can utilize the saved memory to train a larger backbone (Row 3), leading to higher performance while still using less memory. Conventional: conventional non-reversible finetuning. } 
\setlength{\tabcolsep}{2.0pt}
\small
\begin{tabular}{ll|cca}
\toprule
 {Finetuning} & Backbone & \multicolumn{1}{c}{Top-1 acc } & \multicolumn{1}{c}{Top-5 acc } & \multicolumn{1}{a}{Mem (GB)} \\
\hline
Conventional & ViT-small           & 55.2\% & 82.8\%  & 34.2\\
\hline
\textbf{Dr$^2$Net}    & ViT-small   &    65.0\% &88.9\% &8.0 \\
\textbf{Dr$^2$Net}    & ViT-base  & \textbf{68.2\%}   & \textbf{90.8\%}  &  16.6 \\
\bottomrule
\end{tabular}\label{tab:acc_vs_memory_dinov2}
\end{table}

\begin{table}
\centering
\caption{\textbf{Accuracy versus memory for action recognition~\cite{goyal2017something} with VideoMAE~\cite{tong2022videomae} pretrained models.} Our Dr$^2$Net can utilize the saved memory to train a larger backbone (Row 3), leading to higher performance while still using less memory. Conventional: conventional non-reversible finetuning. } 
\setlength{\tabcolsep}{2.0pt}
\small
\begin{tabular}{ll|cca}
\toprule
 {Finetuning} & Backbone & \multicolumn{1}{c}{Top-1 acc} & \multicolumn{1}{c}{Top-5 acc} & \multicolumn{1}{a}{Mem (GB)} \\
\hline
Conventional & ViT-small           & 66.5\% & 90.3\%     &   29.3    \\
\hline
\textbf{Dr$^2$Net}    & ViT-small   &  64.6\% & {89.0\%}   &  6.0  \\
\textbf{Dr$^2$Net}    & ViT-base  &  \textbf{68.6\%} & \textbf{92.0\%}      & 13.0 \\
\bottomrule
\end{tabular}\label{tab:acc_vs_memory_videomae}
\end{table}

\begin{table}
\centering
\caption{\textbf{Accuracy versus memory for object detection~\cite{zhang2023dino}.}   Our Dr$^2$Net can utilize the saved memory to train a larger backbone (Row 3\&4), leading to higher performance while still using less memory. Conventional: conventional non-reversible finetuning. Conventional: conventional non-reversible finetuning. } 
\setlength{\tabcolsep}{2.0pt}
\small
\begin{tabular}{ll|ca}
\toprule
 {Finetuning} & Backbone & \multicolumn{1}{c}{AP (\%)} & \multicolumn{1}{a}{Mem (GB)} \\
\hline
Conventional          & Vswin-tiny    & 51.3 & 54.0  \\
\hline
\textbf{Dr$^2$Net}    & Vswin-tiny    & 51.3 & 30.0  \\
\textbf{Dr$^2$Net}    & Vswin-small   & 52.8  & 30.1 \\
\textbf{Dr$^2$Net}    & Vswin-base    & \textbf{54.7}  & 32.4 \\
\bottomrule
\end{tabular}\label{tab:acc_vs_memory_dinov2_odetection}
\end{table}


{
    \small
    \bibliographystyle{ieeenat_fullname}
    \bibliography{main}
}

\end{document}